# NsBM-GAT: A Non-stationary Block Maximum and Graph Attention Framework for General Traffic Crash Risk Prediction

Kequan Chen, Pan Liu, Yuxuan Wang, David Z. W. Wang, Yifan Dai, Zhibin Li

***Abstract*—Accurate prediction of traffic crash risks for individual vehicles is essential for enhancing vehicle safety, as it forms the foundation for designing and implementing proactive crash prevention strategies. While significant attention has been given to traffic crash risk prediction, existing studies face two main challenges: First, due to the scarcity of individual vehicle data before crashes, most models rely on hypothetical scenarios deemed dangerous by researchers. This raises doubts about their applicability to actual pre-crash conditions. Second, some crash risk prediction frameworks were learned from dashcam videos. Although such videos capture the pre-crash behavior of individual vehicles, they often lack critical information about the movements of surrounding vehicles. However, the interaction between a vehicle and its surrounding vehicles is highly influential in crash occurrences. To overcome these challenges, we develop a novel non-stationary extreme value theory (EVT), where the covariate function is optimized in a nonlinear fashion using a graph attention network. The EVT component incorporates the stochastic nature of crashes through probability distribution, which enhances model interpretability. Notably, the nonlinear covariate function enables the model to capture the interactive behavior between the target vehicle and its multiple surrounding vehicles, facilitating crash risk prediction across different driving tasks. We train and test our model using 100 sets of vehicle trajectory data before real crashes, collected via drones over three years from merging and weaving segments. We demonstrate that our model successfully learns micro-level precursors of crashes and fits a more accurate distribution with the aid of the nonlinear covariate function. Our experiments on the testing dataset show that the proposed model outperforms existing models by providing more accurate predictions for both rear-end and sideswipe crashes simultaneously.**

This work was supported in part by National Natural Science Foundation of China (51925801, 52232012, 52272331) *(Corresponding author: Pan Liu)*

Kequan Chen is with the School of Transportation, Southeast University, China, also is with the School of Civil and Environmental Engineering, Nanyang Technological University, Singapore 639798(e-mail: Chenkequan@seu.edu.cn)

Pan Liu and Zhibin Li are with the School of Transportation, Southeast University, China (e-mail: Liupan@seu.edu.cn; lizhibin@seu.edu.cn)

Yuxuan Wang is with the State Key Laboratory of Internet of Things for Smart City, University of Macau, Macau, China (e-mail: yxwang@um.edu.mo)

David Z. W. Wang is with the School of Civil and Environmental Engineering, Nanyang Technological University, Singapore 639798 (e-mail: wangzhiwei@ntu.edu.sg).

Yifan Dai is with the Tsinghua University Suzhou Automotive Research Institute, Suzhou, China (e-mail: daiyifan@tsari.tsinghua.edu.cn)

## I. INTRODUCTION

Road safety is a critical global issue. Approximately 1.19 million people die each year due to road traffic crashes [1]. With continuous advancements in onboard equipment and detection methods, proactive traffic control systems have emerged as a promising approach to enhancing road safety [2]. In these systems, crash risk prediction plays a crucial role in guiding both the formulation and effectiveness of proactive strategies.

Currently, there is growing interest in developing crash risk prediction models [3], [4], [5], [6], [7]. Some recent studies have explored deep learning approaches to enhance prediction accuracy [8], [9], [10]. However, most of these works rely on aggregated data collected between 15 minutes and 30 seconds before a crash. While we acknowledge that each driver's behavior is unique, such aggregated data fails to capture these individual differences, which limits the accuracy of predictions for individual vehicles. Moreover, such data is not well suited for predicting crashes in very short time frames, such as 3s or even shorter, prior to the crash moment. Dashcam data presents a promising alternative for anticipating crashes over shorter intervals [11], [12]. Yet, since dashcams typically capture only the front view, they fail to record information from all surrounding vehicles. As a result, these studies are mainly applicable to simple two-vehicle involved crash scenarios and struggle to provide accurate prediction results in complex driving tasks, such as merging, diverging, or weaving scenarios. Additionally, some of these studies only focus on crash risk associated with other vehicles captured by dashcams, neglecting the safety of the subject vehicle itself.

In this study, we investigate ways to provide a general crash risk prediction for individual vehicles across different driving tasks. Our proposed model, **NsBM-GAT** (Non-stationary Block Maximum Graph Attention Network), is designed and learned from the real-world vehicle trajectory data prior to crashes to achieve this goal. **NsBM-GAT** consists of two main components: **Non-stationary Block Maxima (NsBM)** and **Graph Attention Network (GAT)**.

*NsBM*: Since traffic crashes have a long-tail characteristic compared to normal traffic scenarios, collecting large enough and diverse vehicle interaction data before crashes is quite challenging [13]. To address the issue of limited crash data, we construct the NsBM framework. NsBM enables the generation of generalized extreme value (GEV) distributions for different danger levels by using extremely dangerous events rather than all events [7]. A key challenge with the NsBM framework is the need to predefine what constitutes an extremely dangerous event. Existing studies typically rely on one or two measurements to identify such events for constructing the GEV distributions [14], [15], [16]. However, vehicles are often affected by approximately six surrounding vehicles during their driving tasks [17]. Thus, simple measurements may not fully capture the complexity of these events. Even though some studies use covariates functions to incorporate additional vehicle interaction information, most of these functions are linear and fall short of modeling the inherent complexity of driving tasks.

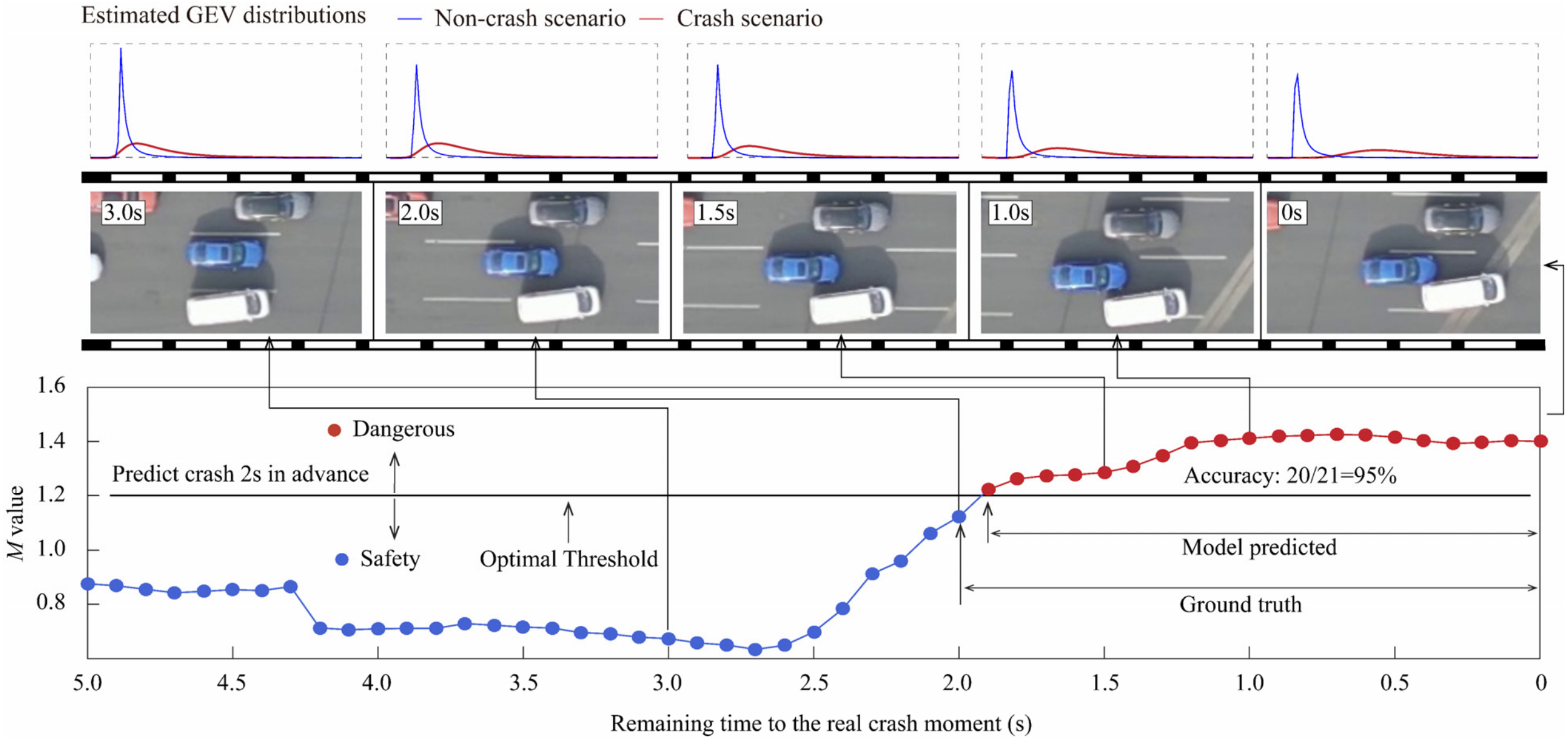

Fig. 1. Overview of our proposed method applied on a real lane-changing crash sample collected by the drone. The time interval is 0.1s. In the context of predicting crash 2s in advance, our proposed NsBM-GAT model first estimated two GEV distributions for each timestep based on the subject vehicle (blue car) and its relationships with surrounding vehicles. The blue distribution corresponds to the non-crash state, while the red distribution represents the crash state. A larger divergence between these distributions indicates a higher crash risk, quantified by a value denoted by M. The black solid line represents the optimal threshold for predicting a crash 2s in advance. When the model's output exceeds the optimal threshold, a warning is issued for the current state. The red circle represents the warning moments proposed by our model (1.9s before the crash moment). Out of 21 timesteps within the 2 s pre-crash interval, 20 are correctly predicted, yielding an accuracy of 95% (20/21). Please visit our website for more detailed demos (https://github.com/KeqC/NsBM_GAT.git).

*GAT*: Inspired by graph attention networks (GATs), which have been widely used to predict vehicle trajectories and traffic flow in road networks [18], [19], [20], [21], we model the relationship between the target vehicle and its six surrounding vehicles as a graph. In this graph, nodes represent the vehicles, and the edges represent their interactions. One of the main contributions of this paper is applying GATs to render the covariate functions in the NsBM model nonlinear. This enhancement facilitates the construction of a general crash risk prediction model that is applicable to diverse driving tasks. The loss function of GATs is now designed as the negative log-likelihood of the simulated GEV distribution, with the objective of minimizing negative log-likelihood to improve the fitting performance.

Vehicle trajectory data before a crash is a valuable resource for developing real-time crash risk prediction models. However, due to the inherent randomness and rarity of crashes, obtaining complete vehicle trajectory data before a crash is extremely challenging. To overcome this issue, we conducted a three-year drone video recording project in merging and weaving segments in Nanjing, China. For the first time, we captured the entire process of 100 crashes, both before and after they occurred. We use the crash trajectory data from the merging segment to train our proposed NsBM-GAT model. The training effectiveness is evaluated using probability-probability (P-P) plots and continuous ranked probability score (CRPS). We also compare these fitting performances with a baseline model. In addition, the rear-end and sideswipe crashes from the weaving segment are used to assess the predictive performance of NsBM-GAT. Experiment results show that the GEV distributions fitted by our approach are closer to the true data distribution and yield better predictive performance on different driving tasks. Figure 1 illustrates the prediction process of our model prior to crash occurrence. This demo is available on our website (https://github.com/KeqC/NsBM_GAT.git). The main contribution of this paper can be summarized as follows:

1. We introduce the first general crash risk prediction model for individual vehicles, which leverages vehicle trajectory data recorded prior to various crash types.

2. We develop a novel graph attention network to convert the linear covariate function in the non-stationary EVT model into a nonlinear one. This design captures intricate interdependence among vehicles while maintaining theoretical interpretability.

3. We demonstrate that the distribution obtained using our model aligns better with the real-world distribution. We demonstrate that our model can achieve better crash risk prediction under different crash types in different datasets, as an example shown in Figure 1.

4. Our model employs a graph structure to capture the interactions between the target vehicle and its surrounding vehicles. Consequently, it shows strong potential for integration with Bird's Eye View (BEV) or occupancy-based autonomous driving systems for crash risk assessment.

## II. Related Works

### A. Crash Risk Prediction Models

Traditionally, research related to crash risk prediction models begins by classifying data collected from loop detectors into crash and non-crash conditions. Then, statistical or deep learning models are used to analyze the compiled data to establish the relationships between hazardous traffic conditions and the likelihood of a crash. For example, in [5] the authors integrated aggregated traffic data from loop detectors with connected vehicle data collected 5 to 10 minutes before crashes. They employed a bidirectional long short-term memory (LSTM) model to predict crash potential, achieving promising results. Cai et al [8] proposed a deep convolutional generative adversarial network to analyze the aggregated traffic data leading to crashes. Their model generated synthetic crash data that better matched the real data distribution, improving prediction performance compared to other methods. Man et al [10] proposed GAN and transfer learning to investigate the aggregated information collected 1 to 6 min before a crash. The extensive coverage of loop detectors enables continuous collection of traffic crash data, which provides robust support for the development of a real-time crash prediction model. However, due to the sparse placement of loop detectors, only aggregated traffic metrics, such as average speed, average flow, and average density, can be captured. As a result, the above-mentioned models are suitable for assessing crash likelihood at a regional level, but they are limited in predicting crash risk for individual vehicles.

Dashcam cameras record videos before the occurrence of crashes, providing detailed vehicle motion data for crash detection and anticipation research. This field has garnered significant attention in recent years, as it holds the potential to enhance the safety of autonomous vehicles [11], [12], [22], [23], [24], [25], [26], [27], [28]. The main challenge they aim to tackle is how to address the rapid motion of multiple vehicles in dashcam videos, dense traffic scenes, and limited visual cues. Commonly used public dashcam video datasets include DADA-2000 [27], AnAn Accident Detection (A3D) [28], Dashcam Accident Dataset (DAD) [29], and Car Crash Dataset (CCD) [30]. Fang et al [27] introduced the DADA-2000 dataset, a large-scale benchmark with 2000 video clips owning about 658, 476 frames covering 54 crashes. They focused on linking driver attention, including eye fixations and scan paths, with crash prediction to enhance driving safety. Fatima [26] combined feature aggregation (FA) and LSTM to anticipate crashes from dashcam video sequences. In [30], graph convolutional networks (GCNs) and Bayesian deep learning were applied to incorporate spatial and temporal features of vehicle interactions. Bao et al [11] further developed Deep Reinforced crash anticipation with Visual Explanation (DRIVE) to address the lack of visual explanation in existing methods. Yao et al [24] proposed an unsupervised network to detect crashes by predicting the future location of vehicles. In real complex driving environments, such as a busy merging segment on a highway, a potential crash with the preceding vehicle may be caused not only by direct interactions but also by the influence of surrounding vehicles [31]. However, dashcam video typically fails to capture comprehensive information about all the surrounding vehicles, which poses limitations for real-time crash prediction in more intricate driving scenarios.

### B. Graph Networks

Graph networks have been widely applied in transportation research. For example, traffic road networks can naturally be represented as graphs, where intersections serve as nodes and road segments as edges. This representation has enabled significant advancements in network-wide traffic forecasting through graph-based approaches. For example, Bogaerts et al [32] developed a Graph CNN-LSTM hybrid model for traffic condition prediction by utilizing GPS data from DiDi's ride-hailing services in the cities of Xi'an and Chengdu, China. Their architecture captured spatial dependencies through graph convolutions and temporal dynamics via LSTM modules. Gui et al [33] further advanced this field by introducing a graph wavelet gated recurrent (GWGR) neural network that incorporates spatial features through wavelet transforms. Graph network in [34] considers both static factors and dynamic factors (e.g., spatial distance, semantic distance, road characteristic, road situation, and global context) to predict traffic flow. In [35] the authors introduced a Multi-relational Synchronous Graph Attention Network (MS-GAT), which learns complex traffic data couplings through traffic data-based channel, temporal, and spatial relations between nodes.

In recent years, graph neural networks (GNNs) have also attracted increasing attention in the field of vehicle trajectory prediction [36], [36], [37], [38], [39], [40], [41], [42], [43]. In this context, each vehicle is represented as a node with its corresponding attributes as node features, and the interactions between nodes form the edges. This graph-based approach captures richer information compared to traditional physics-based prediction methods. For example, Yang et al [36] proposed a GNN-based architecture to capture both temporal and spatial dependencies among vehicles. In addition to the historical trajectory of the subject vehicle, Chen et al [41] emphasized that the historical trajectories of surrounding vehicles are crucial for accurate prediction. They introduced the Sparse Attention Graph Convolution Network (SAGCN) to address this need. GCN-based networks in [37] and [42] have demonstrated the capability to predict the trajectories of both the subject vehicle and its surrounding vehicles simultaneously.

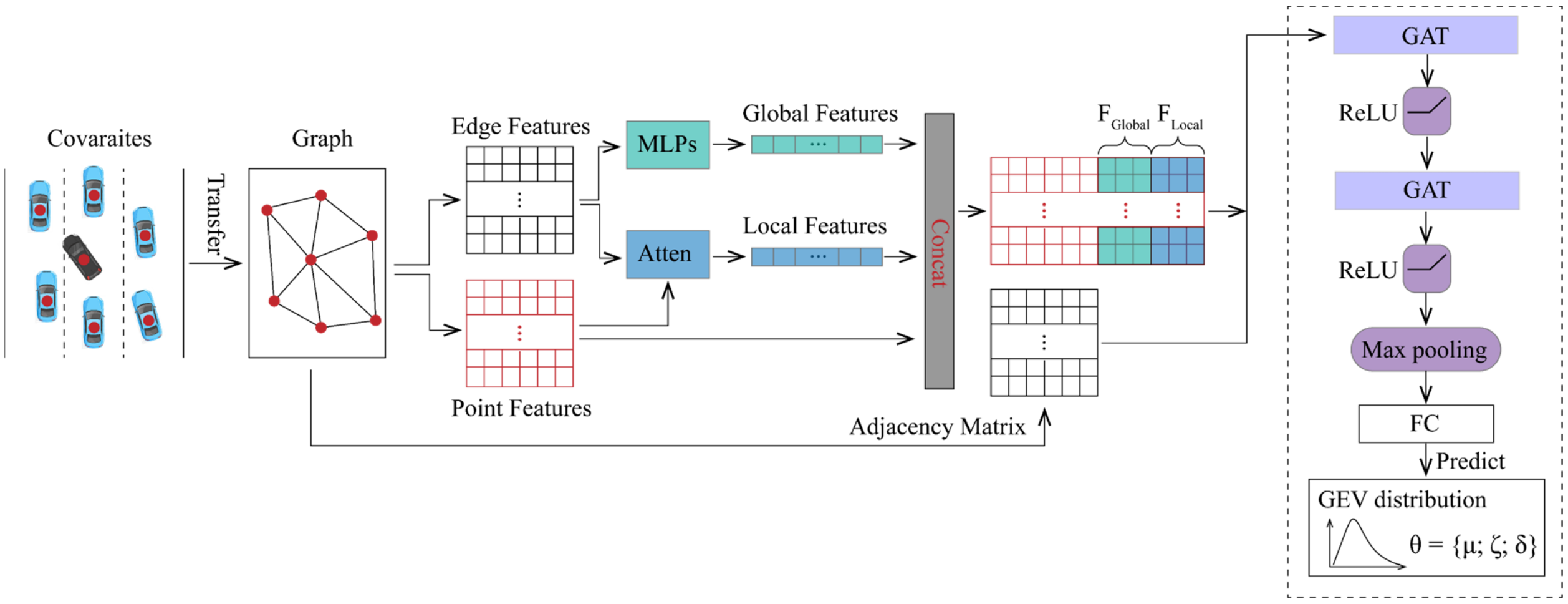

Fig. 2. Overview of the proposed NsBM-GAT framework. Our NsBM-GAT learns distribution patterns from graph structures corresponding to pre-crash scenarios. MLP and attention layers integrate the weights of edge features. Then two GAT layers followed by a fully connected layer predict GEV distribution parameters for each graph structure.

### C. Extreme Value Theory

Extreme Value Theory (EVT) is a statistical framework used to analyze the probability distribution of rare tail events. It has been widely applied in fields such as climate science [44], financial risk management [45], and engineering reliability analysis [46]. Given that traffic crashes are critical long-tail phenomena, EVT-based crash frequency prediction has emerged as a significant research area [14], [47], [48], [49]. The earliest concrete EVT model for linking extremely dangerous events to traffic crashes was developed by Songchitruksa and Tarko [50]. The core idea is to fit asymptotic extreme value distributions to observed dangerous events, thereby enabling extrapolation beyond recorded extremes to estimate crash probabilities. This approach rigorously bridges the gap between observable dangerous events and rare crash occurrences.

Current studies commonly employ surrogate safety measures (SSMs) (e.g., Time-to-Collision (TTC), post encroachment time (PET), and Deceleration Rate to Avoid Crash (DRAC)) to evaluate the risk level across all traffic events. Then, extremely dangerous events will be selected by a predefined threshold of SSMs for subsequent extreme value distribution modeling. Two principal EVT approaches dominate this field: Block Maxima (BM) and Peaks Over Threshold (POT). The BM approach constructs Generalized Extreme Value (GEV) distribution by analyzing maxima within time-series observations, while the POT approach develops Generalized Pareto (GPD) distributions through threshold-exceeding analysis of cross-sectional samples. Considering the fact that vehicle trajectories are also time-series data, several recent studies have attempted to use the BM approach to predict crashes in real time for individual vehicles [7], [51]. For example, Chen et al [7] extracted extremely dangerous events from 5s trajectories data before the crash moment across 20 crash samples and constructed corresponding GEV distributions. Crucially, parallel GEV distribution was developed for non-crash scenarios under identical operational conditions. The differential of these GEV distributions enables real-time crash prediction.

In typical traffic scenarios, vehicle interactions exhibit hexagonal spatial dependencies. Specifically, a subject vehicle's operational state is simultaneously influenced by six adjacent vehicles (front/rear pairs in left, original, and right lanes). This multi-agent interdependency creates nonlinear coupling effects that traditionally SSMs fail to quantify. Recent studies have attempted to address this issue by adding a covariate function into the BM approach [52], [53], [54]. For example, Ali et al [52] considered the impact of speed, spacing, remaining distance, and lag gap to develop a non-stationary BM model for the crash estimation of the mandatory lane-changing (LC) maneuver.

While current covariate formulations rely on linear operators, they inherently prevent the ability to predict crash risks across multiple driving tasks simultaneously. For example, a unified linear covariates formulation is insufficient to capture the distinct risk factors influencing both car-following and lane-changing maneuvers. Furthermore, even when analyzing a single driving task, the linear covariate function is not enough to capture the intricate interactions among vehicles.

## III. The Proposed Method

In this section, we introduce the proposed Non-stationary Block Maximum Graph Attention Network (NsBM-GAT) framework for general crash risk prediction at the individual vehicle level. Figure 2 illustrates the architecture of our model. Before detailing the model construction, we first describe its basic settings.

Our model learns from real-world vehicle trajectories prior to crashes (both rear-end and sideswipe crashes). The model works for individual vehicles at each time step. When predicting the occurrence of a crash, the timing of the prediction is crucial. For instance, in an extreme case, if we only aim to predict a crash 0.1s in advance, this is relatively easy to achieve because the remaining distance between crash-involved vehicles at this moment is nearly zero. This phenomenon is typically present in crash scenarios and not in non-crash scenarios. Thus, the prediction for 0.1s advance of a crash would almost be 100% accurate, with no false positives. However, such a prediction is of limited use for proactive crash prevention, as it only leaves 0.1s for emergency response strategies. In contrast, predicting a crash further in advance tends to reduce the warning accuracy. Therefore, one key challenge addressed in this paper is the prediction of crashes at varying lead times. In the following discussion, we use $T$ to denote the lead time prior to a crash. For different lead times, the trajectory data regarding the crash samples, spanning from time 0 (crash moment) to $T$ is used to train our model.

Specifically, we first present the traditional Block Maximum (BM) method to address the distribution fitting problem for extremely dangerous events in Section III-A. In Section III-B, we address the limitations of linear covariate formulations in the BM method by proposing a novel solution based on Graph Attention Networks (GAT). Then, we describe the design of the loss function and gradient computation for integrating the BM and GAT approaches in Section III-C. Finally, we introduce the real-time crash risk prediction method based on NsBM-GAT in Section III-D.

For a given crash sample $i$ and a lead time $T$, the extremely dangerous events are quantified as a time series vector $X^i$:

$$X^i = \left[ X_1^i, X_2^i, \cdots X_t^i \right] \tag{1}$$

where $t = T/\tau$, $\tau$ denotes the time interval (0.1s in this work).

According to a block size $w$ of equal size, we divide the vector $X^i$ into $k$ non-overlapping blocks. The maximum value from each block is retained to form a new sequence $M_k^i$:

$$M_k^i = \max_{(k-1)w<j<kw} X^i \tag{2}$$

Following the approach in [7] and [55], an initial threshold $Q$ is applied to eliminate non-dangerous samples, further refining the sequence $Y^i$ :

$$Y^i = \left\{ M_j^i \mid M_j^i \in \left( M_1^i, M_2^i, \cdots M_k^i \right), M_j^i > Q \right\} \tag{3}$$

For all crash samples, extremely dangerous event sequences $Y^i$ are obtained using (1) to (3), yielding the following:

$$Y = \left\{ Y^1, Y^2, \cdots, Y^n \right\} \tag{4}$$

where $n$ is the total number of crash samples.

Assuming that $Y$ follows a GEV distribution, its cumulative distribution function (CDF) is given as follows:

$$F(z;\theta) = \begin{cases} exp\left\{ -\left[ 1 + \xi \dfrac{z-\mu}{\sigma} \right]^{\frac{-1}{\xi}} \right\}, \xi \neq 0 \\ exp\left\{ -exp\left[ -\dfrac{z-u}{\sigma} \right] \right\}, \xi = 0 \end{cases} \quad \theta = (\xi, \mu, \sigma) \tag{5}$$

where $\xi$ is the shape parameter controlling the tail behavior of the GEV distribution. When $\xi$ equals zero, the distribution corresponds to the Gumbel distribution. $\sigma$ is the scale parameter, determining the spread of the GEV distribution. A larger value of $\sigma$ indicates more variability in the extreme events. $\mu$ is the location parameter, indicating where extreme events tend to occur within the observed data.

*B Graph attention networks*

The interaction behavior between the target vehicle (see the black vehicle in Figure 2) and its surrounding six vehicles (see the blue vehicles in Figure 2) will be the focus of this study. We quantify extremely dangerous events in (1) using the minimum remaining distance (MRD) between the subject vehicle and its neighbors. MRD is chosen because it has a direct physical relationship with crashes. When this value decreases to zero, it corresponds to the occurrence of a crash. Existing research has also used MRD to assess the safety performance of autonomous vehicles [56]. However, MRD has notable limitations. For instance, even with a small MRD, if vehicles are moving in opposite directions, an immediate danger might not exist. The interaction between vehicles can significantly influence the extreme events quantified by MRD. Thus, in this section, we aim to modify the traditional BM approach by incorporating nonlinear graph attention networks. This modification better accounts for the complex effects of vehicle interactions on crash formation.

Regarding the extremely dangerous events selected by (4), we derive their corresponding graph structures, denoted by $G_p$=($V_p$,$E_p$) , where $p$=$\mathcal{N}(r)$ and $\mathcal{N}(r)$ represents all the extremely dangerous events in (4). $V_p$ denotes the set of nodes in extremely dangerous event $p$. Each node $v \in V_p$ is associated with a feature vector $h_v \in \mathrm{R}^{d_v}$, where $d_v$ is the dimensionality of the node features, including the instantaneous speed, acceleration, direction of the movement, lane number, and vehicle position. $E_p$ denotes the set of edges in the extremely dangerous event $p$. As shown in Figure 2, a complete graph structure comprises 11 edges. The features of these edges are represented by a sparse matrix $A$ of size $\mathrm{R}^{E_p \times d_e}$,where $d_e$ is the feature dimensionality of each edge. The edge features include the relative speed and acceleration between vehicles, lateral and longitudinal distance difference, and relative steering angle.

To capture global features from all the edge attributes in the graph structure, we first apply an attention-based multilayer perceptron (MLP) to aggregate and transform the edge features.

$$F_g = AttentionMLPs(A) \tag{6}$$

where $F_g \in \mathbb{R}^{d_g}$, and $d_g$ is the dimensionality of the global features.

Given that the edges connected to nodes have varying levels of importance, we apply an attention mechanism to weigh these edges. The nodes connected to $v$ are represented by $u$, with the edge features denoted by $A_{vu}$. The attention weight between node $v$ and $u$, denoted as $\alpha_{vu}$, is given by:

$$\alpha_{vu} = \frac{exp\left(\text{Leaky ReLU}\left(a^T\left[W_e A_{vu} \parallel W_h h_v\right]\right)\right)}{\sum_{u' \in \mathcal{N}_v(u)} exp\left(\text{Leaky ReLU}\left(a^T\left[W_e A_{vu'} \parallel W_h h_v\right]\right)\right)} \quad (7)$$

where $W_e$ and $W_h$ are the weight matrix associated with the edge and node features, respectively. $a$ is the attention mechanism's parameter vector, and $\mathcal{N}_v(u)$ is the set of nodes connected to node $v$.

Then, we use these weights to aggregate the edge features. The local feature of node $v$, denoted as $F_{L,v}$, is given by:

$$F_{L,v} = \sum_{u \in \mathcal{N}_v(u)} \alpha_{vu} \cdot W_e A_{vu} \quad (8)$$

The original feature of each node $v$ is then concatenated with the local and global features obtained from (6) and (8), respectively. Let $H_v$ denote the updated node representation, which is given as follows:

$$H_v = Concat\left(h_v, F_G, F_{L,v}\right) \quad (9)$$

The GAT model considers both the node itself and its neighboring nodes to compute the attention-based mechanism for updating its information. The updated node representation, obtained from (9), is then used as the input to the GAT networks. The attention mechanism is calculated as follows:

$$e_{vu}^{(l)} = \text{Leaky ReLU}\left(a^T\left[W_G^{(l)} H_v^{(l)} \parallel W_G^{(l)} H_u^{(l)}\right] + b^{(l)}\right) \quad (10)$$

$$\alpha_{vu}^{(l)} = \frac{exp\left(e_{vu}^{(l)}\right)}{\sum_{u' \in \mathcal{N}_v(u)} exp\left(e_{vu'}^{(l)}\right)} \quad (11)$$

where $e_{vn}^{(l)}$ is the attention coefficient of node u with respect to node $v$ at the $l$-th layer, $\alpha_{vn}^{(l)}$ is the attention coefficient of $e_{vn}^{(l)}$ after applying the softmax function, $W_G^{(l)}$ and $b^{(l)}$ are the weight and bias terms at the $l$-th layer of the GAT, $a^T$ is the transpose of the attention parameters, which are used to compute the attention score between node features, the symbol || is the join operation of matrix, and the LeakyReLU function is employed to introduce non-linearity into the model.

The attention coefficients computed from (11) are then used to calculate the output of each GAT network for each node:

$$H_v^{(l+1)} = \text{ReLU}\left(\sum_{u \in \mathcal{N}_v(u)} \alpha_{vu}^{(l)} W_G^{(l)} H_v^{(l)} + b^l\right), \; for\; l = 1,2 \quad (12)$$

The output from the second layer of the GAT is passed through a max pooling operation. This process effectively reduces the dimensionality of the node features by aggregating the most prominent features across the graph, which can be represented as:

$$H_{pooled} = \text{MaxPooling}\left(H_v^{(3)}, \forall v \in V_p\right) \quad (13)$$

Finally, the vector $H_{\text{pooled}}$ is passed to a fully connected (FC) layer. The FC layer learns the mapping from the pooled feature representation to the three raw parameters of the GEV distribution, which are the raw shape parameter $\tilde{\xi}_p$, raw scale parameter $\tilde{\sigma}_p$, and raw location parameter $\tilde{\mu}_p$. Considering the constraints of the GEV distribution, these raw parameters require further transformations.

For the shape parameter, we apply the hyperbolic tangent function to constrain its value between -1 and 1, ensuring it is within the valid range for the GEV distribution. Then, we have:

$$\xi_p = tanh\left(\tilde{\xi}_p\right) \quad (14)$$

Given that the scale parameter must be strictly positive in the GEV distribution, we apply the exponential function to ensure the output is always greater than zero. Then, we have:

$$\sigma_p = exp\left(\tilde{\sigma}_p\right) \quad (15)$$

The location parameter does not require any constraints as it can take any real value, then, $\mu_p = \tilde{\mu}_p$.

*C: Design of loss function*

Once the NsBM-GAT is constructed, it is trained to capture vehicle interaction information from the graph in crash scenarios, with the goal of improving the fit of the predicted GEV distribution to the empirical GEV distribution. For each graph structure used to train the model, three predicted parameters of the GEV distribution can be used to compute the log-likelihood value, denoted as $\log f(z_p)$:

$$\log f\left(z_p\right) = \begin{cases} -\left(1+\frac{1}{\xi_p}\right)\log\left[1+t_p\right] - \left[1+t_p\right]^{-\frac{1}{\xi_p}} - \log\sigma_p, \; \xi_p \neq 0 \\ -\left[\frac{z_p-\mu_p}{\sigma_p}\right] - \exp\left\{\left[\frac{z_p-\mu_p}{\sigma_p}\right]\right\} - \log\sigma_p, \; \xi_p = 0 \end{cases} \quad (16)$$

$$t_p = \frac{\xi_p\left(z_p-\mu_p\right)}{\sigma_p} \quad (17)$$

Regarding a batch of training data, we compute the sum of the negative log-likelihoods, denoted as $\mathcal{L}$, to guide the model's training. The goal is to minimize $\mathcal{L}$ using gradient descent.

$$\mathcal{L} = -\sum_{p \in B(p)} \log f\left(z_p \mid \mu_p, \sigma_p, \xi_p\right) \quad (18)$$

Here, we detail the gradients of the loss function with respect to each parameter of the GEV distribution. When $\xi_p \neq 0$, the gradients for the $\xi_p$, $\sigma_p$, and $\mu_p$ can be calculated using the corresponding derivatives of the negative log-likelihood function, which are shown in the (19)-(21).

$$\frac{\partial\mathcal{L}}{\partial\xi_p} = \frac{1}{\xi_p^2}\log s_p - \frac{t_p}{\xi_p s_p} + \frac{s_p^{-\frac{1}{\xi_p}}}{\xi_p^2}\left(\log s_p - \frac{\xi_p t_p}{s_p}\right) \quad (19)$$

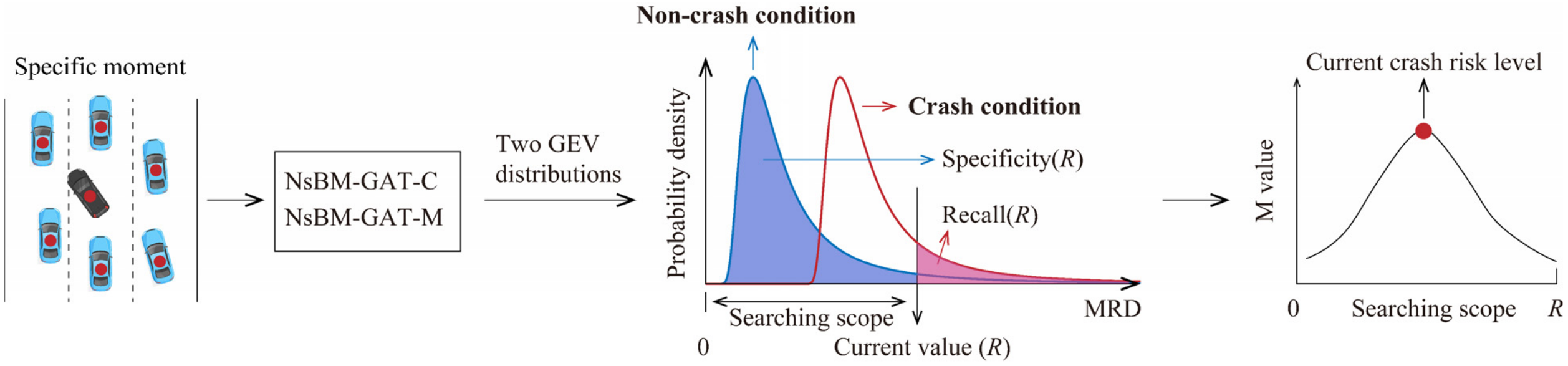

Fig. 3. Crash risk prediction using NsBM-GAT based GEV distributions

$$\frac{\partial \mathcal{L}}{\partial \sigma_p} = \frac{1}{\sigma_p}\left[-1+\left(\frac{1}{\xi_p}+1\right)\frac{\xi_p t_p}{s_p} - \frac{\xi_p t_p}{\sigma_p} s_p^{-\frac{1}{\xi_p}-1}\right] \tag{20}$$

$$\frac{\partial \mathcal{L}}{\partial \mu_p} = \frac{1}{\sigma_p}\left[\left(\frac{1}{\xi_p}+1\right)\frac{\xi_p}{s_p} - \frac{\xi_p}{\sigma_p} s_p^{-\frac{1}{\xi_p}-1}\right] \tag{21}$$

$$s_p = 1+\xi_p t_p \tag{22}$$

When $\xi_p = 0$, (19)-(21) are simplified as:

$$\frac{\partial \mathcal{L}}{\partial \mu_p} = \frac{1}{\sigma_p}\left(1-e^{-t_p}\right), \frac{\partial \mathcal{L}}{\partial \sigma_p} = \left(-1+t_p - t_p e^{-t_p}\right), \frac{\partial \mathcal{L}}{\partial \xi_p} = 0 \tag{23}$$

Due to transformations applied to the scale and shape parameters after applying the FC layer, their gradients need to be adjusted as follows:

$$\frac{\partial \mathcal{L}}{\partial \tilde{\sigma}_p} = \frac{\partial \mathcal{L}}{\partial \sigma_p}\cdot \sigma_p \tag{24}$$

$$\frac{\partial \mathcal{L}}{\partial \tilde{\xi}_p} = \frac{\partial \mathcal{L}}{\partial \xi_p}\cdot\frac{\partial \xi_p}{\partial \tilde{\xi}_p} = \frac{\partial \mathcal{L}}{\partial \xi_p}\cdot\left(1-\tanh^2\left(\tilde{\xi}_p\right)\right) \tag{25}$$

These above-mentioned gradients are then used to update the weights and biases of NsBM-GAT through gradient descent:

$$\Theta_{N,new} = \Theta_{N,old} - \eta \sum_{k\{\mu_p,\sigma_p,\xi_p\}} \frac{\partial \mathcal{L}}{\partial k}\frac{\partial k}{\partial \theta} \tag{26}$$

where Θ represents the model parameters, $\eta$ is the learning rate

*D Real-time crash risk prediction method*

The framework proposed in Sections III-A and III-B can also be applied to extremely dangerous events observed in non-crash conditions. Consequently, we construct two separate networks: NsBM-GAT-C for crash scenarios and NsBM-GAT-N for non-crash scenarios. For any given graph structure, these networks generate GEV distributions, enabling real-time prediction. Specifically, if the current state closely resembles a crash scenario, it should have a higher probability of appearing in the crash GEV distribution and a lower probability in the non-crash GEV distribution. Based on this assumption, a real-time crash risk prediction model is built by comparing the differences between the two predicted GEV distributions.

Figure 3 illustrates the process of crash risk prediction using the NsBM-GAT framework. For a given time step, we first estimate two GEV distributions $F_{\text{crash}}(x)$ and $F_{\text{non-crash}}(x)$. With a given MRD value (denoted as $R$), the area above $R$ in the $F_{\text{crash}}(x)$ represents the probability of correctly identifying a crash in crash samples, which is referred to as Recall($R$). Conversely, the area below $R$ in the $F_{\text{non-crash}}(x)$ represents the probability of correctly identifying a non-crash in non-crash samples, known as Specificity($R$). A value of $R$ that achieves both high Recall and Specificity indicates that the current state exhibits clear differentiation between crash and non-crash scenarios. Thus, we use (27) to evaluate the crash prediction ability for a given MRD value [7]:

$$Metric(R) = Recall(R) + Specificity(R) \tag{27}$$

Following (27), we assess the prediction capability for all MRD thresholds from 0 to the current value ($R$), as shown by the block curve in Figure 3. The max value of this curve can be used to quantify the crash risk level, denoted by $M$:

$$M = \max_{0<i<R}\left(Metric(i)\right) \tag{28}$$

The range of the $M$ value is from 0 to 2. A larger $M$ value indicates a higher probability that the current state corresponds to a crash condition, while simultaneously suggesting a lower probability of it being a non-crash condition.

## IV. Experimental Setting

In this section, we first introduce the vehicle trajectory data before crashes, which we use for training and testing our models (Section IV-A). We then describe the evaluation methods employed to assess the fitting performance and prediction accuracy (Section IV-B). Finally, we introduce the baseline models used for comparison (Section IV-C).

*A Crash dataset*

Unlike existing crash report data, loop detector data, or dashcam video data, this study utilizes vehicle trajectory data before crashes collected by drones. This kind of crash data captures high-precision information about all vehicles surrounding the crash vehicle. Given the inherent randomness and rarity of crashes, capturing pre-crash information is particularly challenging. To overcome the difficulty, the research team conducted a three-year drone video recordings project aimed at obtaining high-quality crash data. Specifically, data were collected on two buy expressways in Nanjing, China, which form the basis of our training and testing datasets, as shown in Figure 4.

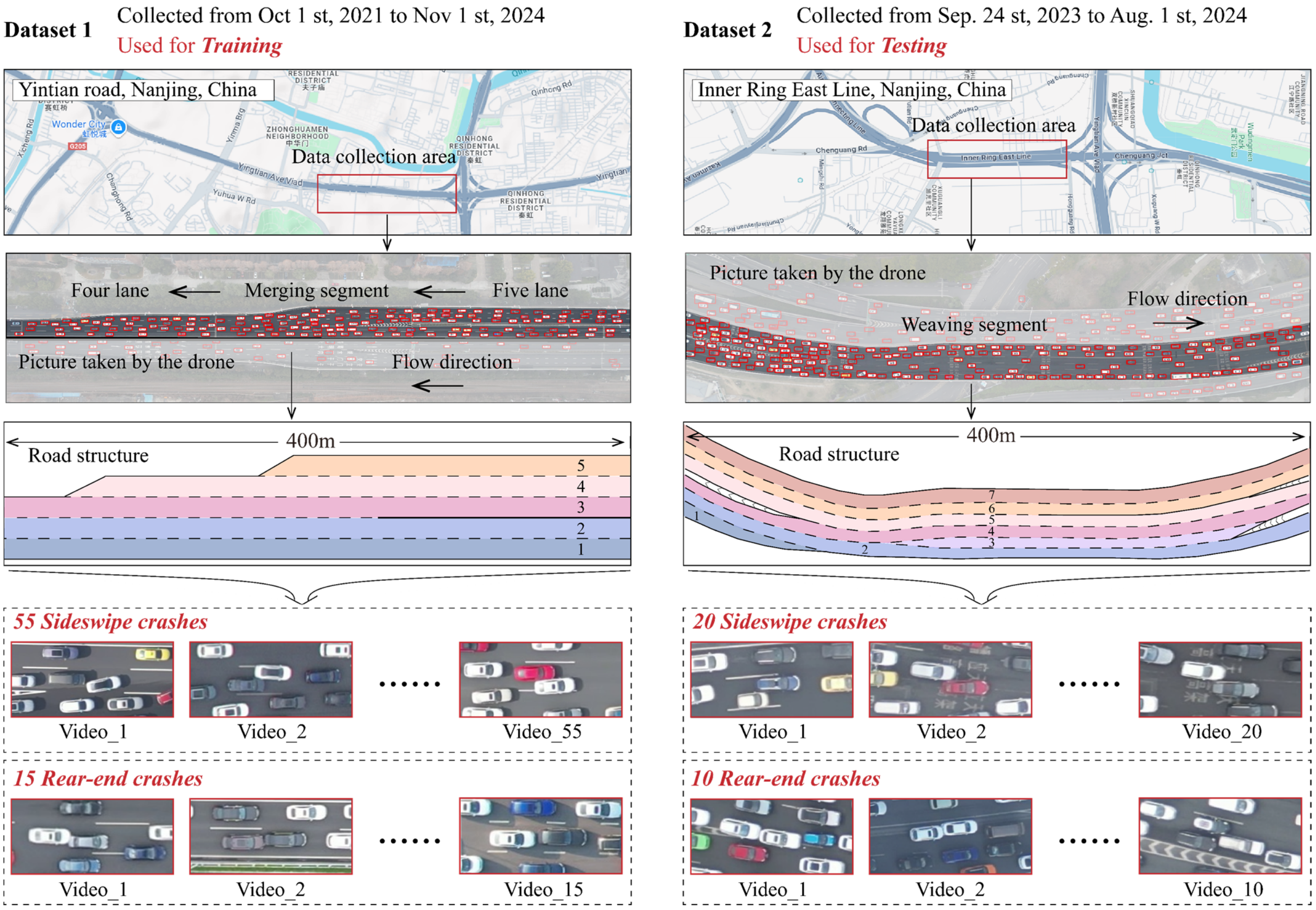

Fig. 4. Real-world crash data recorded by drones. **Left** shows a busy merging segment in Nanjing, China, where 55 sideswipe crashes and 15 rear-end crashes were captured, along with a large amount of non-crash data. This data forms the crash training dataset (**TrDC**) and non-crash training dataset (**TrDN**) for training NsBM-GAT. **Right** shows a busy weaving segment, also in Nanjing, where 20 sideswipe crashes and 10 rear-end crashes were recorded. These two crash types constitute the testing datasets (**TeDC_s** and **TeDC_r**) for validating our model's prediction capability in a different location. Meanwhile, the non-crash data from this site (**TeDN**) is used to evaluate the model's overall performance.

***Training Dataset***: As shown on the left side of Figure 4, a busy merging segment is selected as our study site. Drone video recordings were conducted during the morning peak hours (7:30 AM to 9:30 AM) on every sunny working day from October 1, 2021, to November 1, 2024. The morning peak period was chosen due to the high traffic volume, which increases the likelihood of crashes. The drone was flown at an altitude of 300m, covering approximately 400 meters of roadway. During the 3-year recording period, we recorded 55 sideswipe crashes and 15 rear-end crashes, with detailed information about the before and after of each crash. Using our previously developed vehicle trajectory extraction algorithms [57], [58], [59], we processed the videos to extract high-quality trajectory data. This data is sampled at 0.1s intervals and includes each vehicle's speed, acceleration, moving direction, lane number, and position at each time step. Specifically, the training dataset is divided into two parts: the training dataset for crash scenarios (**TrDC**), and the training dataset for non-crash scenarios (**TrDN**). For the former, we consider the trajectory data of each crash vehicle during the 5s prior to the crash moment. To ensure a fair comparison, non-crash traffic trajectory data from the 5 minutes preceding each crash are selected to form TrDN.

***Testing dataset***: To evaluate the predictive ability of our model, we conducted video recording tasks from September 24, 2023, to August 1, 2024, at another busy weaving segment in Nanjing, China, as shown in the right side of Figure 4. Recordings were performed during the morning peak hours (7:30 AM to 9:30 AM) on every sunny working day. Totally, we recorded 20 sideswipe crashes and 10 rear-end crashes. The trajectory data before and after these crashes are also extracted. Consistent with the classification of the training datasets, we divided the testing dataset into crash scenario (**TeDC**) and non-crash (**TeDN**). To evaluate the performance of our proposed model across different crash types, we divided TeDC into a sideswipe crash testing dataset (**TeDC_s**) and a rear-end crash testing dataset (**TeDC_r**).

*B Evaluation Criteria*

***Fitting performance***: We evaluate the GEV distribution estimated by NsBM-GAT on the training dataset using two methods: the Probability plot (P-P plot) and the Continuous Ranked Probability Score (CRPS). The P-P plot is a common graphical tool to assess the fit of the GEV distribution [14], [48], [54]. To create the P-P plot, we plot the empirical CDF against the theoretical CDF for each sample in the training dataset. If the empirical data perfectly fits the theoretical GEV distribution, the points in the P-P plot should lie along the line $y=x$. The CRPS is a statistical measure that quantifies the difference between CDF and the observed value [60]. Its formula is given as follows:

$$\mathrm{CRPS}_{\mathrm{arg}} = \frac{1}{m}\sum_{m\in\mathcal{N}(r)}\int_{-\infty}^{\infty}\left(F\left(y;\theta_m\right)-1_{y\geq z_m}\right)^2 dy \tag{29}$$

where $\theta_m$ is the predicted three GEV parameters for extreme event m, $1_{y\geqslant z_m}$ is the Heaviside step function (which is 1 when $y\geqslant z_m$ and 0 otherwise). A smaller CRPS indicates a smaller difference between the predicted distribution and the empirical distribution.

***Prediction performance***: We evaluate prediction accuracy for crash samples in the testing dataset using two metrics. First, we use an AP (Average Precision) metric. Previous studies defined AP as a measure of the model's ability to distinguish between crash and non-crash by checking whether the predicted crash time exactly matches a designated moment. If it does, the prediction is successful; otherwise, it is a failure. However, this binary approach neglects the inherent randomness of crashes. In this study, we redefine AP as the proportion of correct predictions within a given advance prediction interval. This approach treats crash prediction as a regression problem rather than a binary classification task. Mathematically, it can be expressed by:

$$AP^T = \frac{\sum_{i=1}^{n}\left(\frac{\sum_{t=0}^{T}\left(\omega_i^t\cdot\tau\right)}{T}\right)}{n} \tag{30}$$

$$\omega_i^t = \begin{cases} 0, if\, M_t < M^* \\ 1, if\, M_t \geq M^* \end{cases} \tag{31}$$

where $\omega_t$ is a binary variable, if the value of $M$ at time t exceeds the optimal threshold $M^*$, then $\omega_t$ is set to 1, otherwise it is set to 0, and $\tau$ represents the time interval, which is set to 0.1s in this study.

To account for false alarms in non-crash samples, we measure overall performance using the area under the receiver operating characteristic (ROC) curve (AUC). Before calculating the AUC, we designate crash samples that satisfy a specified condition $T$ as cases and select controls from non-crash samples in a 1:5 ratio. For instance, when our goal is to predict the crash 1s in advance ($T$ = 1s), all data points within 1s before a crash in the TeDC are treated as crash cases. Using our model, we perform a crash risk assessment for both cases and controls and compute the crash risk value according to (28). By adjusting the threshold of $M$, we obtain recall and false alarm rate values at different thresholds. These values allow us to generate the ROC curve, and the area under this curve is the AUC. A model with a higher AUC demonstrates better performance, as it shows improved discrimination between crash and non-crash samples.

*C Comparison models*

The first comparison model is the traditional BM model without covariates, a simplified version of our proposed model. This model is constructed using only the methods outlined in Section III-A and Section III-D. We call it the stationary BM (SBM) model. Since the SBM model also estimates the GEV distribution, we compare both its fitting and prediction performance with those of our NsBM-GAT model.

Before this study, few crash risk prediction models used trajectory data prior to crashes. Most existing works rely on SSMs to detect potential traffic conflicts and predict crash risk through conflict classification. This approach is common in autonomous driving safety research. To benchmark our model, we also evaluate three widely used SSMs: TTC, MTTC, and DRAC. TTC is a time-based indicator that estimates the time remaining until a potential crash if vehicles continue at their current speed and direction. TTC has a common threshold of 1.5s. Values below this threshold indicate a higher likelihood of crashes [61]. MTTC extends TTC by incorporating vehicle acceleration [62]. Like TTC, its common threshold is also 1.5s. DRAC is an acceleration-based indicator, used in [63] to identify dangerous driving situations. When DRAC exceeds the maximum deceleration acceptable to the driver (typically 3.5m/s$^2$), it signals a high crash risk. It is important to note that these indicators directly provide a crash risk value based on their own assumption rather than a distribution. Therefore, we compare these models only in terms of prediction performance (see Section IV-B).

## V. Model Analysis

This section systematically evaluates the performance of our proposed model through four complementary perspectives. In Section V-A, we present the model's distribution fitting results on the training dataset and compare them with a baseline model. To facilitate crash risk assessment, Section V-B details the optimal threshold selection outcomes for different lead times. Based on this well-trained model and its corresponding optimal threshold, we evaluate the prediction performance on both the TeDC_s and TeDC_r in Section V-C. In addition, we quantify the overall predictive efficacy using TeDN. These results are benchmarked against state-of-the-art approaches. Our experimental results demonstrate that the proposed model achieves improved accuracy in recognizing pre-crash patterns. This enables the model to be applied to non-crash trajectory data for monitoring vehicle operational crash risk. In Section V-D, we illustrate the application of our model in non-crash scenarios to detect potential crash risk patterns.

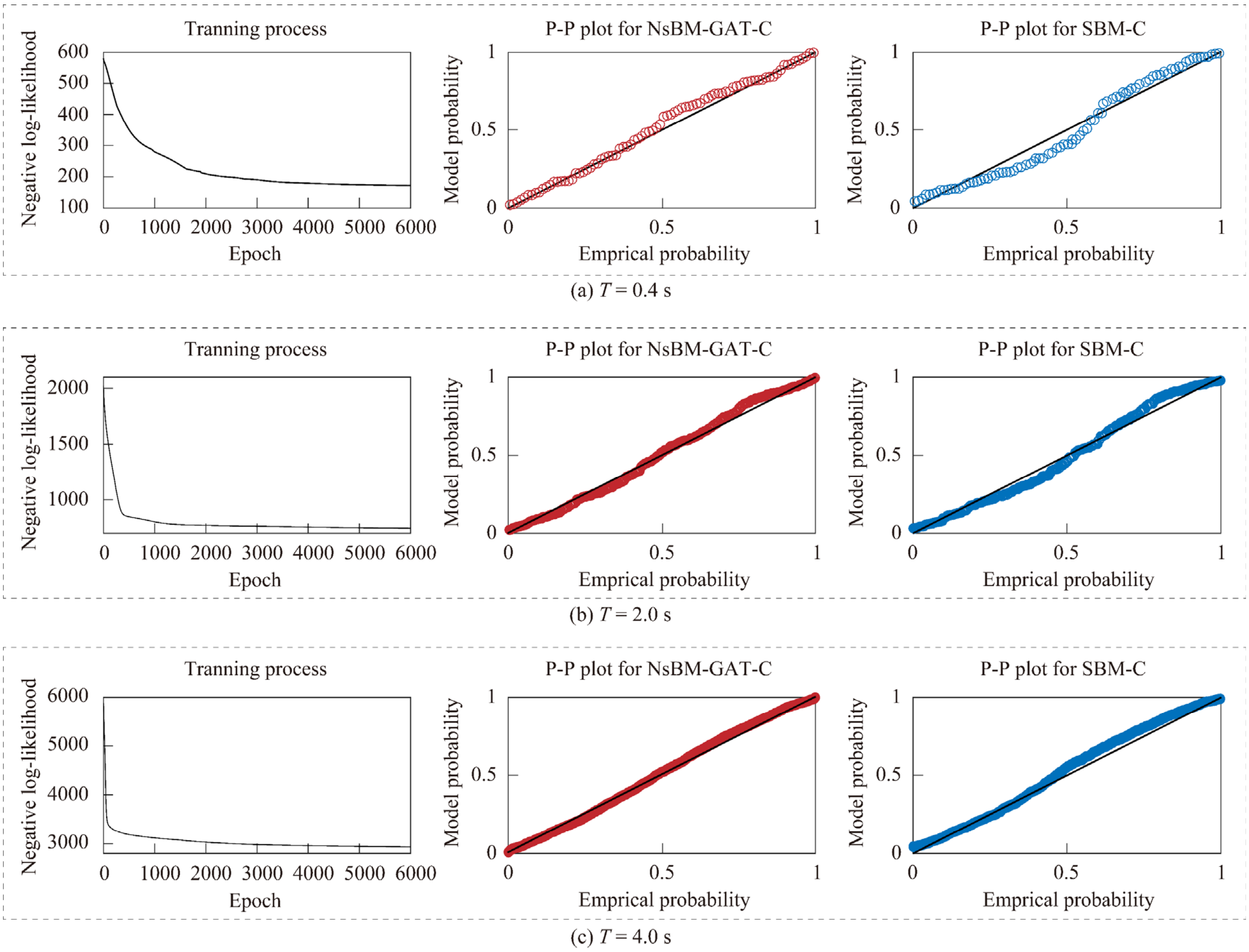

Fig. 5. The training results on the TrDC, including the training process and the P-P plot. The predictions of **ours** are shown in red, while those of **SBM** are in blue.

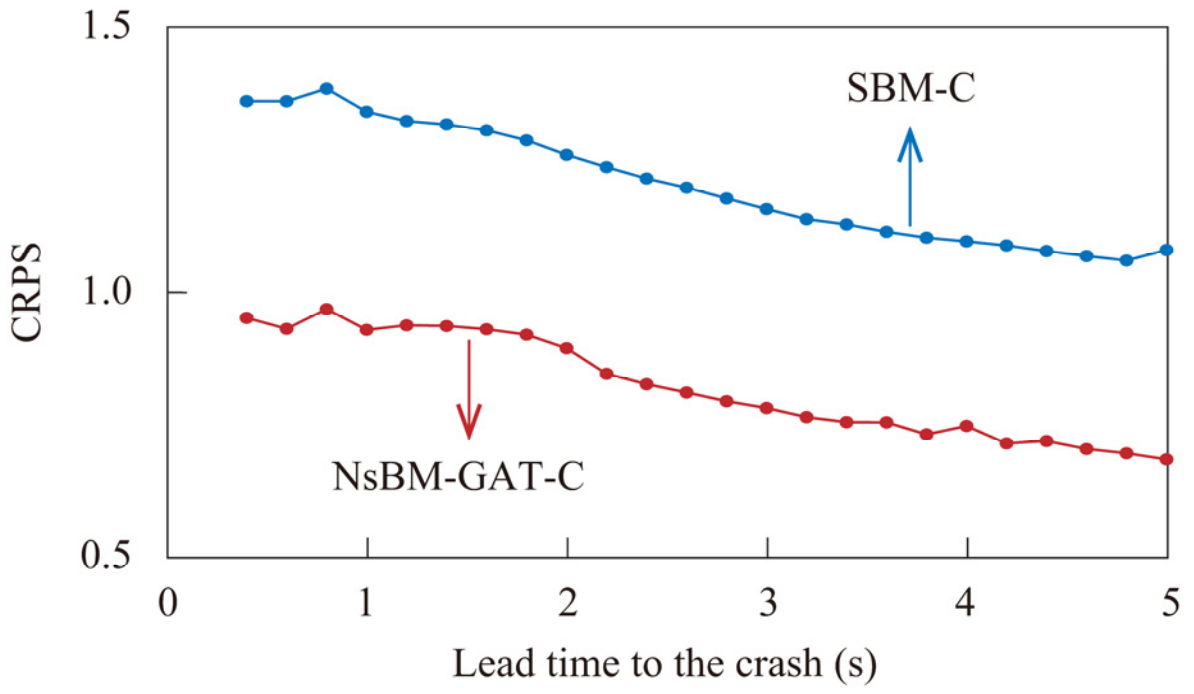

Fig. 6. CRPS comparison of NsBM-GAT-C and SBM-C across different lead times. The proposed model demonstrates superior performance over SBM-C for all tested lead times.

### *A. Training details and results*

To investigate the impact of different lead times ($T$) on model performance, we set the block size in (1) to 0.2s to ensure that each block contains at least two values. The initial threshold in (2) is set to 1m. Under these settings, we perform 24 experiments with $T$ ranging from 0.4s to 5s at 0.2s intervals. The learning rate is set to $10^{-4}$. For example, when predicting a crash 0.4s in advance, we use the graph structure formed by the trajectory data from 70 crashes occurring within 0.4s before the crash as input to train the model. The model outputs the three parameters of the GEV distribution. Figure 5(a) shows the training results of NsBM-GAT-C with $T$ = 0.4s. The negative log-likelihood decreases and stabilizes, indicating that the model is learning effectively. Based on the training data, we compute the P-P plots for NsBM-GAT-C and the baseline model (SBM-C). The P-P plot of our proposed model aligns more closely with the 45° line, which intuitively indicates a better fit of the GEV distribution to the empirical extremely dangerous events within 0.4s prior to crashes. Figures 5(b) and 5(c) show similar results for $T$ = 2s and $T$ = 4s, respectively, further confirming the superior fitting performance of our model.

Next, we calculate the CRPS for different $T$ based on (29), as shown in Figure 6. We find that the CRPS values for NsBM-GAT-C are lower than those for SBM-C across all $T$ values. This result confirms that the GEV distribution fitted by our proposed model is closer to the actual distribution, which is consistent with the intuitive analysis obtained from P-P plots. Regarding the TrDN, we similarly use a 0.2s block size to capture extremely dangerous events for training NsBM-GAT-N and SBM-N. The CRPS values for NsBM-GAT-N and SBM-N are 0.57 and 0.91, respectively. This demonstrates that our model also outperforms SBM-N in fitting extremely dangerous events in the TrDN.

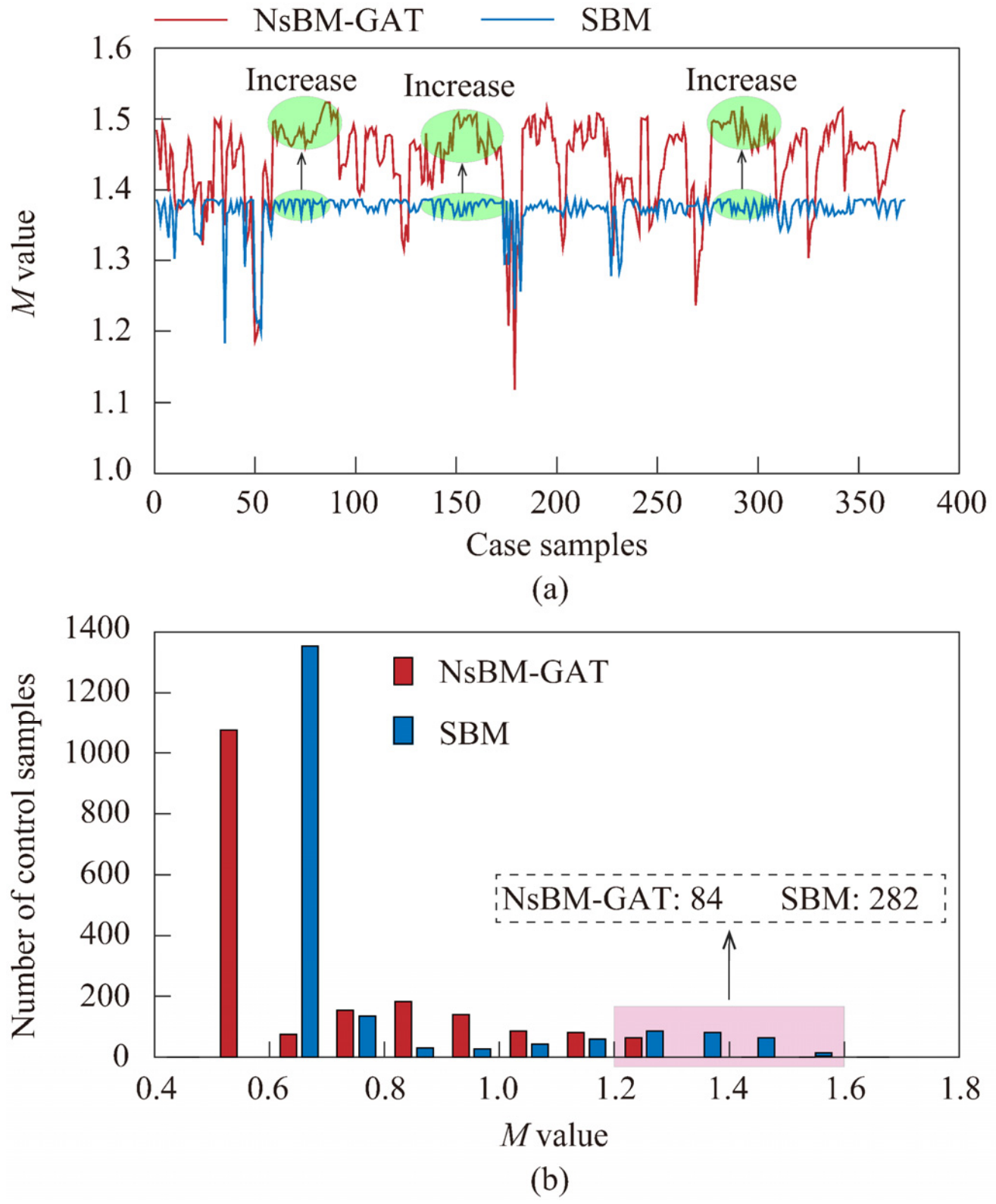

Fig. 7. When T=1s, applying the trained NsBM-GAT and SBM to (a) case samples in TrDC and (b) control samples in TrDN

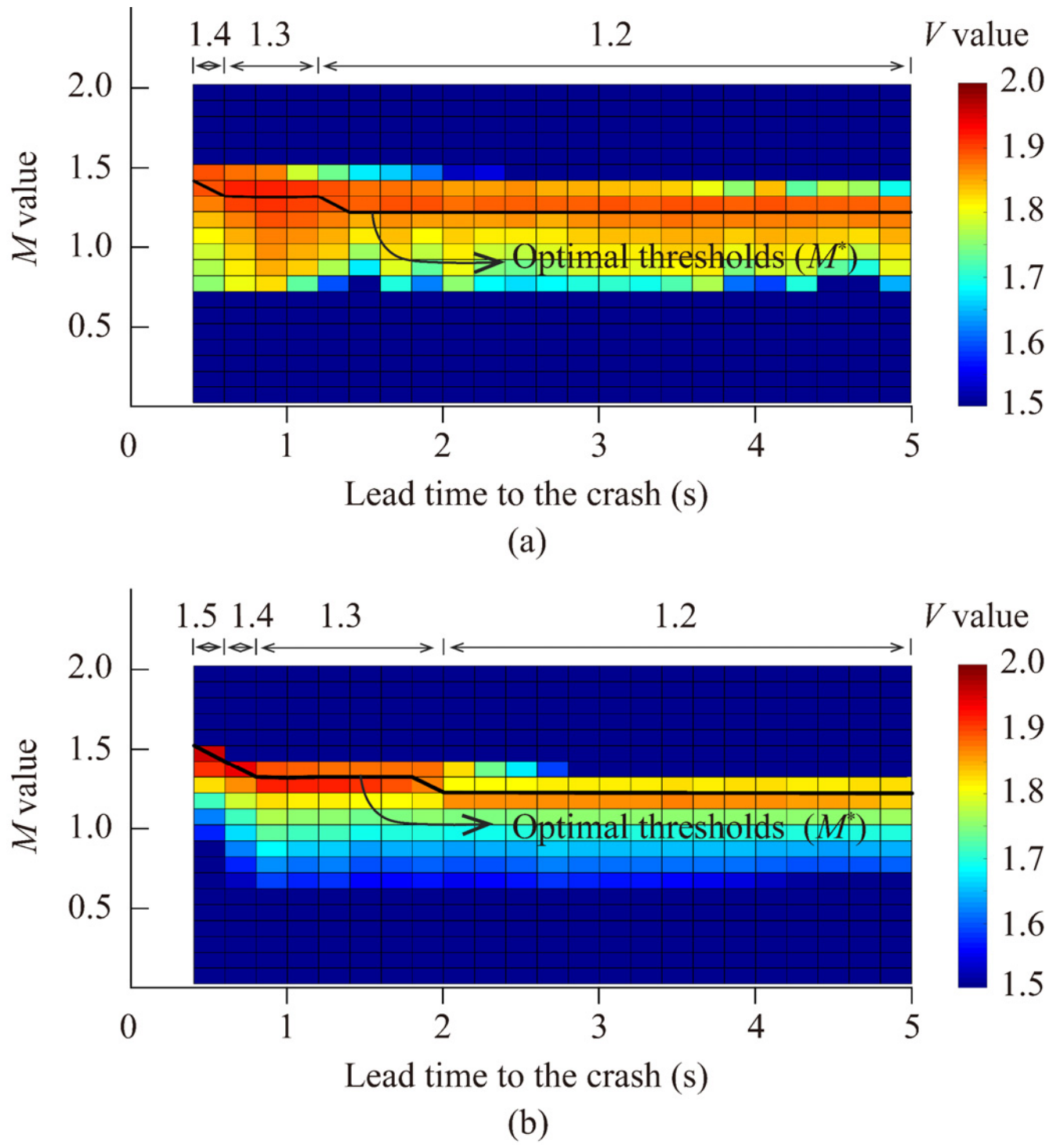

Fig. 8. Determination of the optimal threshold of M across different lead times to the crash ($T$): (a) NsBM-GAT; (b) SBM

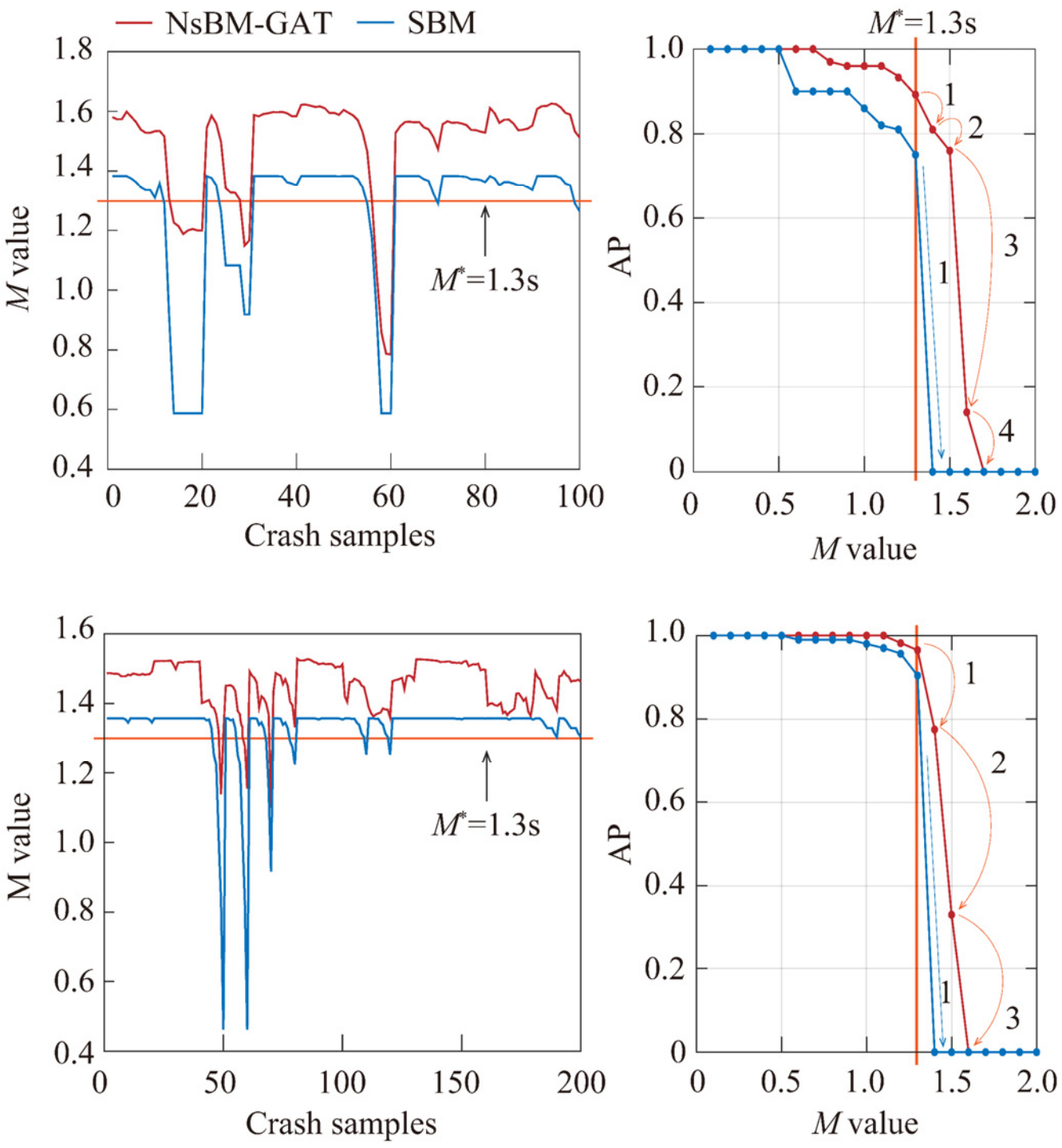

Fig. 9. Crash risk prediction values for the rear-end crash testing dataset (**Top two figures**) and the sideswipe crash testing dataset (**Bottom two figures**) when $T$ = 1s. The red line represents our model, the blue line corresponds to SBM, and the orange line is the optimal threshold determined from the training dataset based on Section V-B. The results are similar to those in Figure 7(a), demonstrating that our model performs robustly on different crash testing samples and achieves higher accuracy compared to the SBM model.

### *B Determination of threshold*

In this section, we determine the threshold of the crash risk prediction model based on the trained NsBM-GAT models. Taking T = 1s as an example, there are a total of 367 inputs from the TrDC. Our well-trained models (NsBM-GAT-C and NsBM-GAT-N) can produce the GEV distribution parameters for each input under both crash and non-crash scenarios. Then, the crash risk is quantified using (27) and (28). The red line in Figure 7 represents the crash risk levels obtained by our proposed model, while the blue line represents the results produced by the SBM model. From this figure, we can see that for most samples, the $M$ values obtained by our model are higher than those from the baseline model. This finding indicates that our model identifies these instances as more dangerous. Given that these samples originate from crash scenarios, higher crash risk values are expected. This result demonstrates that our proposed model can effectively distinguish the distribution patterns of graph structures in crash and non-crash scenarios in the training dataset. Additionally, we select 1835 non-crash samples from TrDN based on a ratio of 1:5 to compute their $M$ values. The results are shown in Figure 7(b). From this figure, we observe that the SBM model tends to estimate larger $M$ values for these non-crash samples. With an $M^*$ of 1.2, the SBM model estimates 282 samples are dangerous, while our model estimates only 84 samples. Since these samples are from non-crash scenarios, high $M$ values are considered false positives.

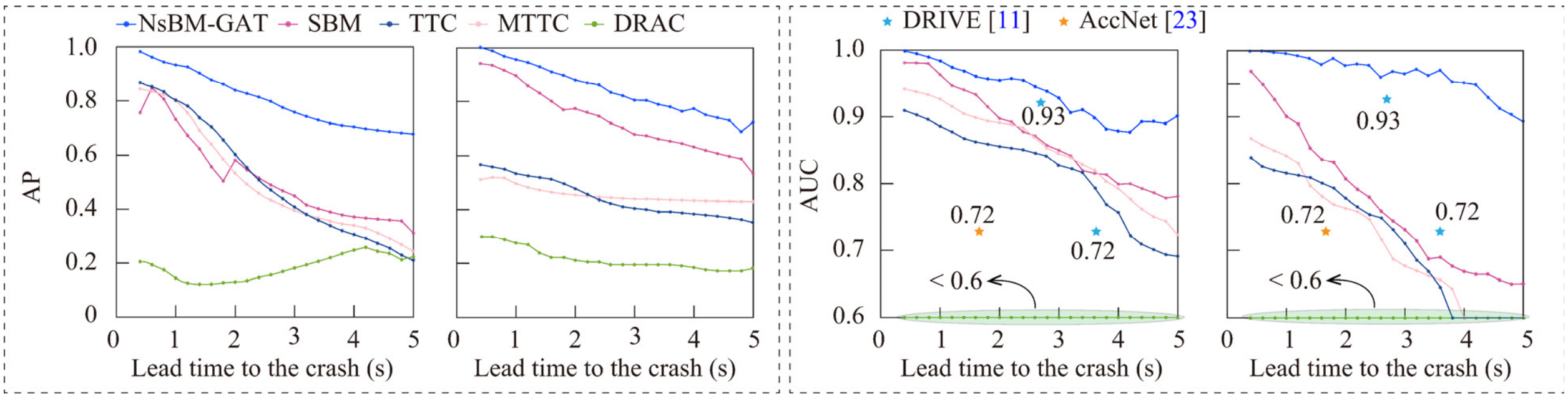

Fig. 10. Prediction performance under various lead times to the crashes. **Left side**: the AP in rear-end crash dataset (left) and sideswipe crash testing dataset (right). **Right side:** the AUC in rear-end crash dataset (left) and sideswipe crash testing dataset (right). The prediction results of SBM and three commonly used methods are compared. Additionally, two AUC results derived from [11] and [23] are included. Across different crash types and lead times, our model consistently outperforms the existing models in predicting crash risk.

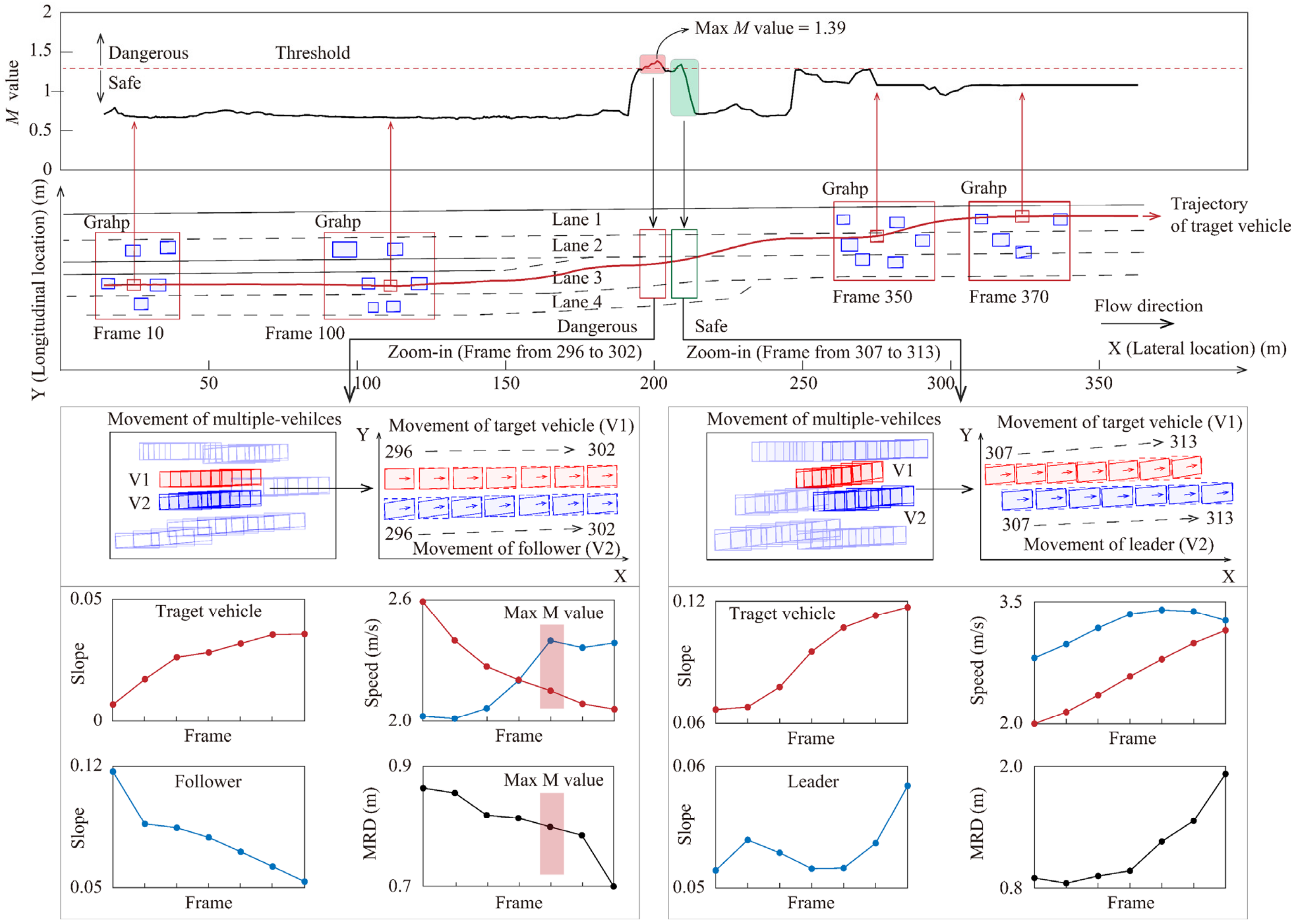

Fig. 11. Application results on real data using our proposed model. The predefined lead time to predict crashes is 1s ($T$=1s). The target vehicle's risk estimation exceeded the threshold from frame 296 to frame 302 and returned to safe levels between frame 307 and frame 313. Detailed plots of the vehicle's movement parameters during these two intervals are presented. The full video of this demo can be found on our website (https://github.com/KeqC/NsBM_GAT.git).

Based on Figure 7, different $M$ thresholds yield distinct Recall and False Alarm Rate (FAR) for crash and non-crash samples. Accordingly, we determine the optimal $M^*$ by maximizing Recall and minimizing FAR, following the condition $V$ = Recall +1 – FAR. We compute the $V$ values for different $M^*$ values at various $T$, with the results for NsBM-GAT and SBM shown in Figure 8(a) and 8(b), respectively. The black lines in Figure 8 represent the optimal $M^*$ for each $T$, where the $V$ value is maximized. For our proposed model, the optimal $M^*$ is 1.4s when $T$ = 0.4s, 1.3s when $T$ ranges from 0.6s to 1.2s, 1.2s when $T$ exceeds 1.2s.

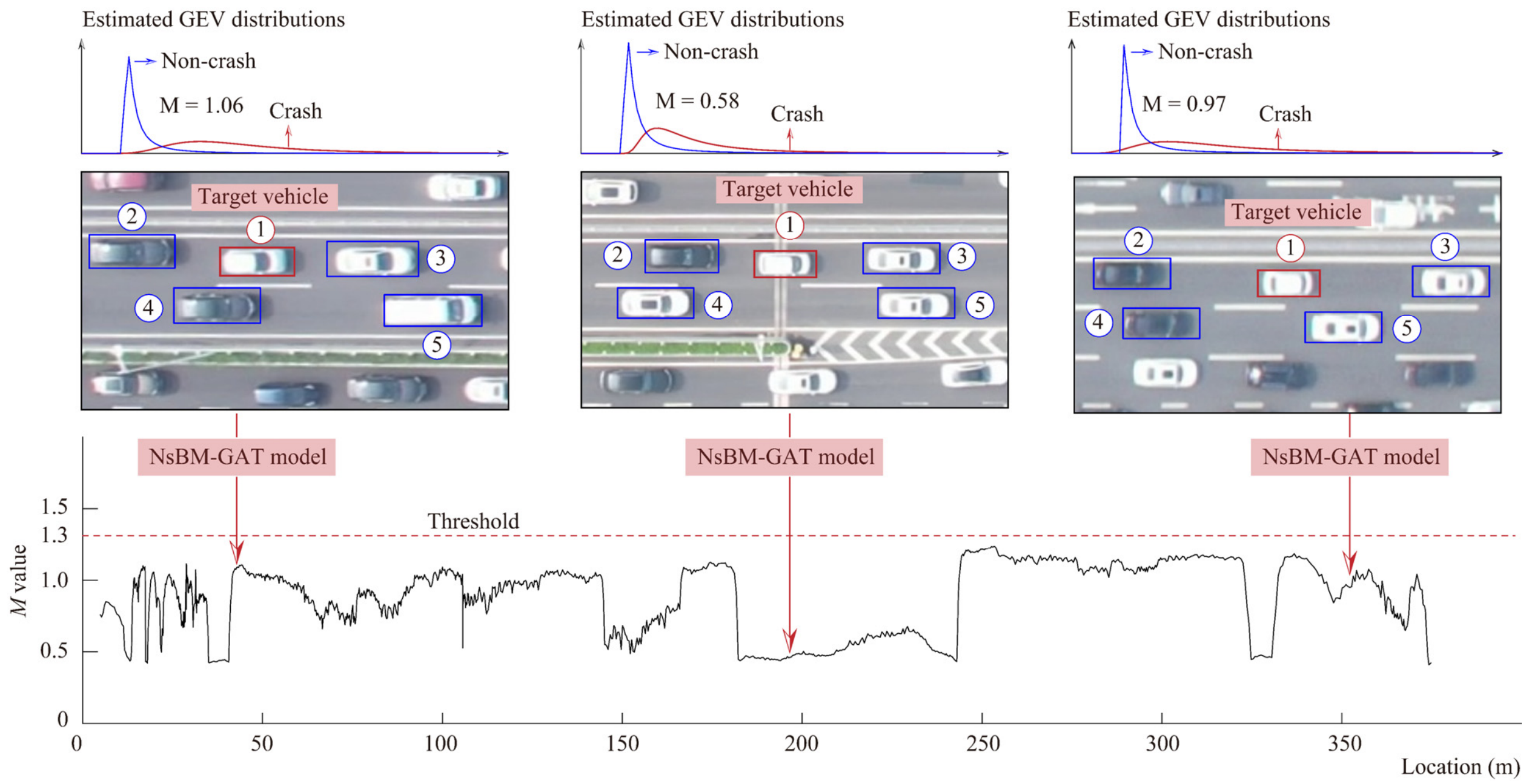

Fig. 12. A car-following scenario with the aim of predicting a crash 1s in advance. The full video of this demo can be found on our website (https://github.com/KeqC/NsBM_GAT.git)

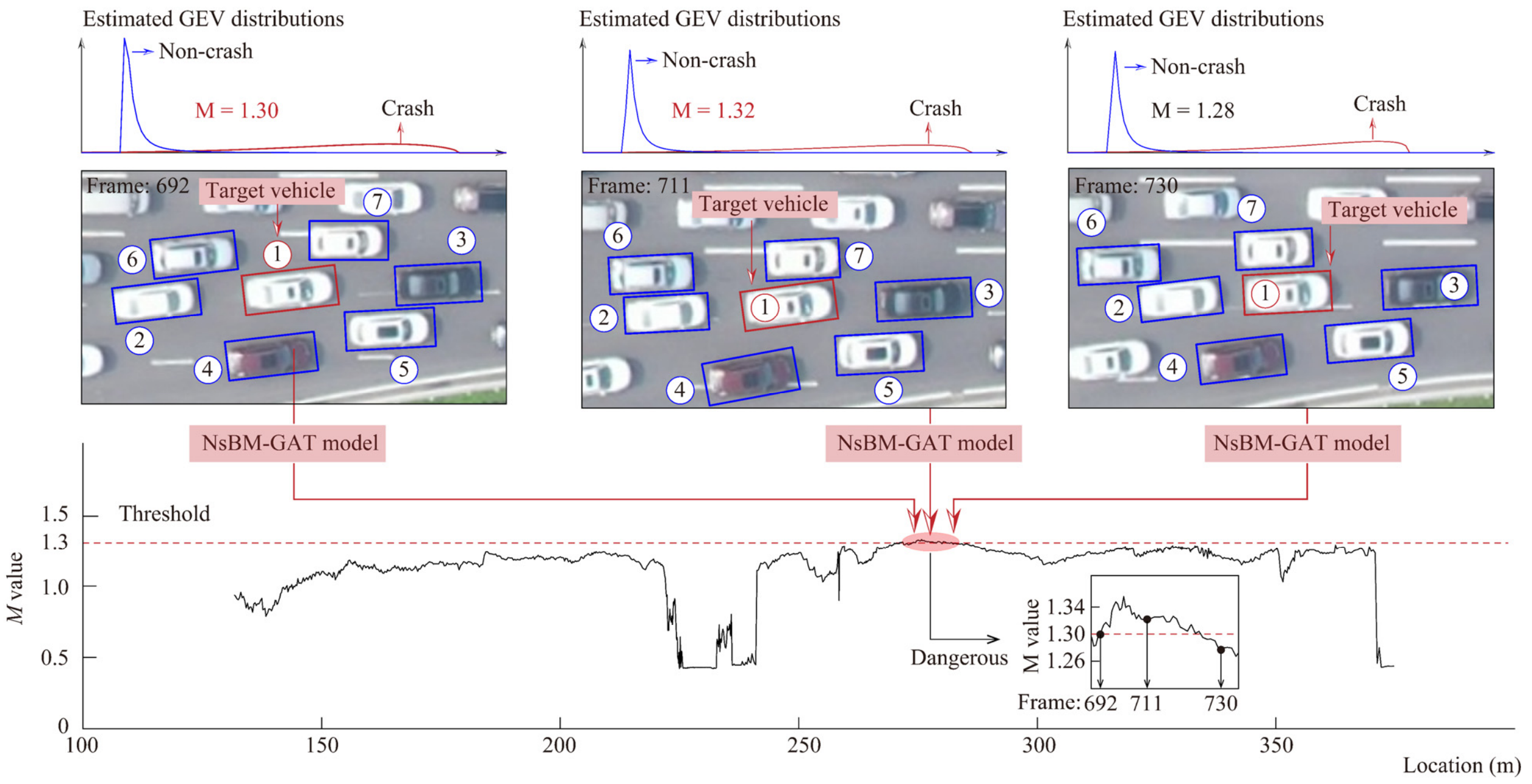

Fig. 13. A lane-changing scenario with the aim of predicting a crash 1s in advance. The full video of this demo can be found on our website (https://github.com/KeqC/NsBM_GAT.git).

*C Prediction performance*

In this section, we evaluate the prediction performance of our proposed model on the TeDC and TeDN using the optimal thresholds. Additionally, SSMs are applied to these datasets to estimate crash risks for comparison.

The red line on the top of Figure 9 shows the estimated $M$ values for samples in the TeDC_r when $T$ =1s using our prediction model. For comparison, the blue line in these figures represents crash risk predictions predicted by SBM. The observed patterns are consistent with those in the training dataset (Figure 7(a)). The $M$ values estimated by our model are significantly higher than those of SBM. This result indicates that the patterns learned from the training dataset are effectively generalized to the rear-end crash testing dataset. The right side of Figure 9 shows the AP, where our model achieves a recall of 89% under the optimal threshold, compared to only 77% for SBM. In addition, the SBM model is highly sensitive to the $M^*$. A 0.1s increase of $M^*$ causes its recall to drop sharply to 0. In contrast, our model maintains a more robust differentiation of dangerous samples, with recall values gradually decreasing as the threshold increases from 1.3s to 1.7s.

For the TeDC_s dataset, we perform a similar AP analysis at $T$ = 1s. The bottom of Figure 9 illustrates the crash risk estimation results for both NsBM-GAT and SBM. We find that NsBM-GAT can predict crash risk for sideswipe crashes with high accuracy, achieving an average AP of 98%. Additionally, NsBM-GAT is less sensitive to threshold changes.

For all values of $T$, we use the AP metric to quantify the model's predictive performance in the TrDC, as shown on the left side of Figure 10. We also calculate the AP values of SSMs at the given threshold, with the results also presented in Figure 10. These results clearly demonstrate that our proposed model significantly outperforms existing methods in terms of AP across various lead times. Notably, as the lead time increases, the AP of all models decreases, primarily due to the reduced predictability of vehicle interactions as the time to the crash increases.

To assess the false positive rate in the non-crash testing dataset, we computed the AUC for our model and the comparison models at different $T$ values, with results shown on the right side of Figure 10. Although existing literature rarely addresses the impact of lead times on model accuracy, both [11] and [23] reported their models' AUC at specific lead times, which we include as comparisons (indicated by the five-point stars in Figure 10). Overall, these results confirm that our proposed model achieves the best AUC compared to existing methods across different driving tasks and lead times.

### *D Applications*

Here, we discuss the application of our proposed model for crash risk prediction. Moreover, we focus on whether the movement patterns of vehicles identified as dangerous in non-crash scenarios by our model exhibit any signs of hazardous driving behavior.

Taking the case of predicting a crash 1s in advance ($T$ = 1s), Figure 11 shows a representative case study. In this case, the target vehicle (V1) gradually changes lanes from lane 3 to lane 1. In the first half of the road, interactions between V1 and surrounding vehicles do not form a dangerous scenario. However, at a location around 200 m, V1's crash risk value abruptly exceeds the threshold (1.3s). The lower part of Figure 11 zooms in on the vehicles' movements between frames 296 and 302. During this interval, the vehicle shown in the blue rectangle (V2) moves towards V1 at a significant angle, while the V1's movement remains stable. This leads to an increase in relative speed and a reduced gap between the two vehicles. Meanwhile, V1 lacks a sufficient gap on its left side for a safety change lens, which elevates the crash risk. Consequently, the scenario formed is quite similar to the situation that 1s before the crash, and the model triggers a warning. Shortly afterward, V1 accelerates leftward as an available gap appears, increasing the distance from V2 and reducing the crash risk below the threshold.

Figure 12 presents a car-following case in which Vehicle 1 consistently follows the vehicle ahead in lane 1. With the aim of predicting the crash 1s in advance, the assessment of our model indicates that Vehicle 1 poses no significant crash risk throughout the process. Drone pictures at the top of Figure 12 imagery confirm that Vehicle 1 maintains a safe distance from surrounding vehicles, and no hazardous behaviors are observed in the video. Moreover, the estimated GEV distributions reveal that at these moments there is a significant overlap, indicating that the states in accident and non-accident scenarios exhibit minimal differences.

Figure 13 presents a lane-change case. During frames 692 to 725, the crash risk value exceeds the threshold. The drone pictures show that Vehicle 1 rapidly approaches Vehicle 7, bringing the two vehicles into close proximity. From the estimated GEV distributions, it can be observed that the two exhibit significant differences, indicating an extremely high probability of a crash. Video analysis reveals that the transition from a dangerous to a safe state occurs when Vehicle 1 changes direction and gradually moves away from Vehicle 7. These video demos (Figures 11 to 13) are available on our website (https://github.com/KeqC/NsBM_GAT.git).

## VI. CONCLUSION

In this paper, we demonstrated the feasibility of learning a general model for predicting crash risk across different driving tasks. We achieved this task by proposing a graph-based extreme value theory framework called NsBM-GAT. In our proposed model, we introduce a BM sampling method from extreme value theory to analyze the distribution of extremely dangerous events prior to crashes. These extreme events are typically influenced by the complex and nonlinear interactions between the target vehicle and its multiple surrounding vehicles. To capture these interactions, we employ a graph attention network to embed nonlinear covariates into the extreme value model. Finally, we collect high-precision vehicle trajectory data before crashes in merging and weaving segments in the real-world to train and test the model. With these efforts, our experiments confirm that incorporating nonlinear vehicle interactions leads to a better fit of extreme events to the GEV distribution. Our crash data enables the model to capture micro-level precursor features of crashes. Furthermore, we demonstrate that the proposed model outperforms existing models in crash risk prediction for rear-end and sideswipe crashes. Our next goal is to streamline and unify the model, enabling more flexible applications for various data collection methods, such as video, radar point clouds, and LiDAR data.